\documentclass[11pt]{article}

\usepackage[a4paper,margin=2.2cm]{geometry}
\usepackage[T1]{fontenc}
\usepackage{microtype}
\usepackage{graphicx}
\usepackage{booktabs}
\usepackage{multirow}
\usepackage{amsmath}
\usepackage{times}
\usepackage[varvw]{newtxmath}
\usepackage{xcolor}
\usepackage{tikz}
\usetikzlibrary{positioning,arrows.meta,fit,calc,decorations.pathreplacing}
\usepackage[font=small,labelfont=bf]{caption}
\usepackage[numbers,sort&compress]{natbib}
\usepackage{enumitem}
\setlist{itemsep=2pt,topsep=3pt,parsep=0pt}
\usepackage{titlesec}
\titleformat{\section}{\large\bfseries}{\thesection}{0.8em}{}
\titleformat{\subsection}{\normalsize\bfseries}{\thesubsection}{0.8em}{}
\titlespacing*{\section}{0pt}{1.4ex plus .4ex}{0.8ex}
\titlespacing*{\subsection}{0pt}{1.1ex plus .3ex}{0.5ex}
\usepackage{fancyhdr}
\usepackage[hidelinks]{hyperref}
\hypersetup{pdftitle={EDU-Jev: Training System-One Decision Models for Education}, pdfauthor={Jiazheng Li, Yulan He}}
\usepackage[section]{placeins}

\definecolor{accent}{HTML}{C2410C}
\newcommand{\edujev}{EDU-Jev}
\newcommand{\up}[1]{\textsuperscript{#1}}


\begin{document}

\begin{center}
{\LARGE\bfseries EDU-Jev: Training System-One Decision Models\\[2pt] for Education}\\[14pt]
{\large Jiazheng Li \qquad Yulan He}\\[5pt]
{\normalsize King's College London}\\[3pt]
{\normalsize \{jiazheng.li, yulan.he\}@kcl.ac.uk}\\[3pt]
{\small October 2026}
\end{center}
\vspace{4pt}

\begin{abstract}
\noindent
Many automated decisions in education involve choosing from a fixed set of options: assigning a rubric score, checking whether an answer meets a marking criterion, or identifying the first wrong step in a solution. Generative language models are increasingly used to make these decisions,  but their responses require parsing, longer outputs increase inference costs, and confidence expressed in words does not necessarily reflect the probability of being correct.
We present \edujev{}, a family of single-pass (``System One'') decision models at 0.8B, 2B and 4B parameters. Given an assessment
context and a  question with predefined answer options, the model  returns a probability distribution over those options without generating text. A
pointer head scores a marker for each option, ensuring the predictions are restricted to the allowed choices.
Training uses a single supervised stage with soft cross-entropy against the available target label distributions; calibration is assessed empirically.
Across twelve education benchmarks, \edujev{} 4B achieves an average accuracy of
67.6\%, exceeding the strongest of five commercial decision APIs by 9.4 percentage points and the best open 4B decision
model by 18.7 points. On four cross-domain education sets,
it matches the strongest API, which remains ahead on
general decision-making and reasoning tasks. The 2B model processes about 150 decisions per second on a single H100 GPU.
\end{abstract}

\section{Introduction}
\label{r:intro}

Educational assessment involves many small decisions: assigning a short answer 0 to 3 marks under a rubric, rating an essay
on several traits, checking whether a marking point is present or absent, identifying the first incorrect step in a solution,
or judging whether a tutor's reply is appropriate. In most cases, the allowed outcomes are fixed in advance.
Automated scoring
has addressed such decisions for decades~\citep{page1966imminence,attali2006erater,burrows2015asag,ke2019aes}, and its operational
requirements are well established: agreement with human markers, consistency across responses, and confidence estimates reliable enough to route uncertain cases to a person~\citep{williamson2012framework}. Feedback is also most useful when learners receive it
while still working on a task~\citep{black1998assessment,hattie2007feedback,shute2008feedback}. As LLM-based systems increasingly
generate feedback with educators in the loop~\citep{zhao-etal-2025-learnlens},
cost and response time become part of the educational challenge.

Large language models are  often used for these tasks as generative judges that write a rationale followed by a
verdict~\citep{zheng2023judge}. In assessment, this approach includes explainable scoring systems that justify each mark against a
rubric~\citep{li-etal-2023-distilling,li-etal-2024-calibrating,li2025aera,li-etal-2025-aera}. Reasoning models and inference-time
reflection extend this process with additional  deliberation~\citep{wei2022cot,deepseek2025r1,li-etal-2025-two}. For assessment, however, this deliberate,
``System Two''~\citep{kahneman2011thinking} approach has three drawbacks. First, generated outputs require parsing and may fall outside the allowed
options, while their rationales may not faithfully reflect how the decision was reached~\citep{li-etal-2025-drift}. Second, inference cost grows with output length, and reasoning models can generate many tokens even for simple decisions~\citep{chen2024overthinking}. Third, their verbalised confidence does not by itself provide
a calibrated probability~\citep{guo2017calibration}.
Moreover, the benefits of chain-of-thought reasoning are mainly on mathematical and symbolic tasks, with smaller gains
on other tasks~\citep{sprague2025cot}.
Most assessment decisions, by contrast, involve judging a response against a given rubric rather than multi-step symbolic reasoning.

A single-pass (``System One'') decision model offers a direct way to make such judgements. It takes a context and a question with a specified answer type and predefined options, then returns
a probability distribution over those options in one forward pass.
This design restricts outputs to valid choices, avoids autoregressive decoding, and allows probabilities to be trained directly using an objective that encourages calibration.
This typed-decision interface was introduced by TypeSafe AI with its hosted decision model Jev~\citep{typesafe2026jev}, which we evaluate through the
version 1.14 API.  JevBench~\citep{jevbench2026} is an independent public benchmark for such models. We refer to open models that implement the same interface as \emph{Jev-style}
models~\citep{openjevdata,openjev,everlcd,together2026jev,patel2026jevsystemone}. They are trained primarily on general decision data and, as our evaluation
shows, are weak on rubric-based assessment. The challenge is to
improve their assessment capabilities
while retaining valid outputs, efficient inference, and an objective that supports reliable probability estimates.

We introduce \edujev{}, a family of single-pass decision models for educational assessment.
Each model reads an
assessment context, such as the item, rubric, reference answer or key points, student response, solution steps, or tutoring dialogues, and answers a choice,
score, or yes/no question. Instead of the language-model output head, a pointer head reads one hidden state per option and returns a
distribution over exactly the declared options (Section~\ref{r:approach}). What changes relative to general decision models is mainly
the training data: a broad mixture of assessment tasks that provide information a human marker would need, including the question and  its
rubric, reference answer, or key points~\citep{wong-etal-2026-questions}.

\begin{figure}[t!]
\centering
\includegraphics[width=0.9\linewidth]{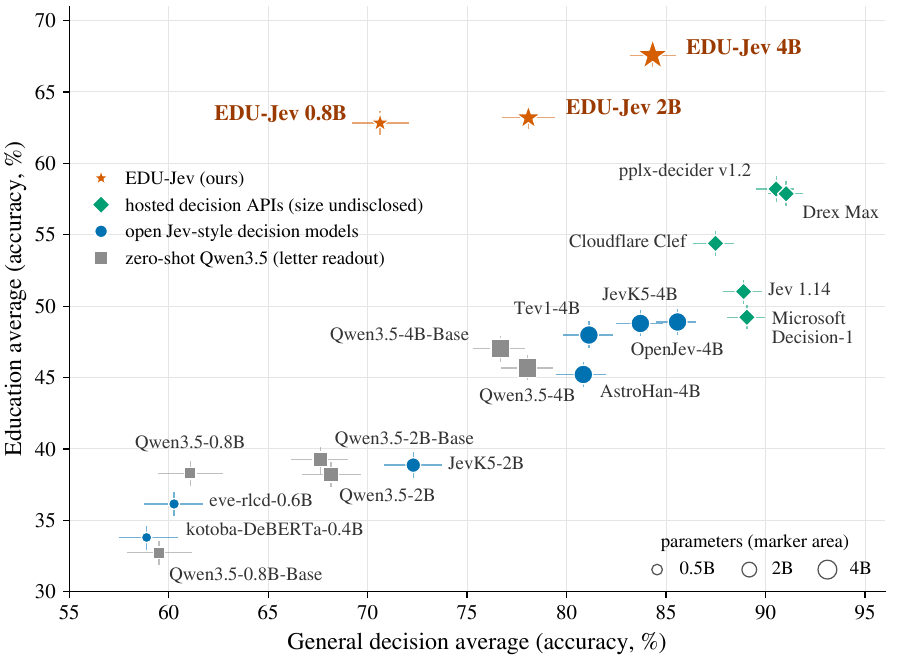}
\caption{Average performance on general versus educational decisions
on the benchmark suite (error bars: 95\% bootstrap confidence intervals; marker area is proportional to parameter
count; hosted APIs, whose parameter counts are undisclosed, use a fixed marker size). \edujev{} models achieve the highest education averages among all
compared models, including hosted APIs. On general decisions, \edujev{} 4B performs similarly to the strongest open 4B models (matching JevK5-4B,
1.3 points below OpenJev-4B) but remains below the APIs.}
\label{r:fig:scatter}
\end{figure}

\edujev{} is trained from Qwen3.5 base models~\citep{qwen35} in a single supervised stage on about 1.5B tokens of typed decisions, with
a strictly proper loss on soft labels and model souping (Section~\ref{r:training}). We evaluate the 0.8B, 2B and 4B models
against open Jev-style models, zero-shot Qwen3.5 models, and five commercial decision APIs on a 20-set suite that separates general
decisions, reasoning, and education within and across domains, with every model scored on the same items
(Sections~\ref{r:eval} and~\ref{r:results}). The 4B model achieves an average accuracy of 67.6\% across the twelve education benchmarks, compared with 45.2--48.9\%
for open 4B decision models and 49.2--58.2\% for commercial APIs (Figure~\ref{r:fig:scatter}).  The 2B (63.2\%) and 0.8B (62.8\%) models also  outperform all evaluated baselines on the education average. On the four cross-domain education sets, the 4B model matches the strongest API. It trails the APIs
by 3.2 to 6.7 percentage points on general decisions and trails three of them by 2.2 to 3.7 percentage points on reasoning.

Our contributions are as follows.
\begin{itemize}
  \item We introduce \edujev{}, a family of 0.8B, 2B and 4B single-pass decision models for education. Their pointer readout restricts predictions to the declared answer options by construction.
  \item We describe a compact supervised recipe and show which choices mattered: broad data coverage, learning-rate decay while training on unseen data,
        and model souping. We also show that reinforcement learning on the pointer readout does not beat continued supervised training, and analyse the reasons for this result (Section~\ref{r:analysis}).
  \item We show that \edujev{} 4B has the highest average accuracy across the education benchmarks among all compared models, exceeding the strongest commercial
        API by 9.4 percentage points.
        In the matched-data comparisons, increasing model size produces larger gains on several transfer and general-reasoning measures than on in-domain assessment accuracy (Sections~\ref{r:res:bench} and~\ref{r:res:scale}).
\end{itemize}

\section{Approach}
\label{r:approach}

\subsection{Typed decisions}
A decision instance is a triple $x=(s,u,\mathcal{O})$. The \emph{state} $s$ is a small set of named text fields: the item, a rubric, a
reference answer or key points, the student response, solution steps, a tutoring dialogue or a passage. Fields that describe provenance
(dataset, split, label source) are never shown to the model. The \emph{question} $u$ has instructions and a type
$\tau\in\{\textsc{choice},\textsc{score},\textsc{yes/no}\}$, and $\mathcal{O}=(o_1,\dots,o_K)$ are its options. The three types share the
input format and the readout and differ only in a type tag in the input and in what the options are: unordered alternatives for
\textsc{choice} ($2\le K\le 255$, e.g.\ the first erroneous step or a misconception), ordered levels with numeric values $v_1<\dots<v_K$
for \textsc{score} ($2\le K\le 10$, e.g.\ 0 to 6 marks or an essay band), and the pair (yes, no) for a binary proposition such as ``the
response makes marking point 3''.

The target is a distribution $q\in\Delta^{K-1}$: the rater histogram when several raters are available, otherwise one-hot. The model
returns $p_\theta(\cdot\mid x)\in\Delta^{K-1}$. From this distribution we derive the predicted option $\hat{k}=\arg\max_k p_k$, a confidence score
$1-H(p)/\log K$, where $H$ is the Shannon entropy, and, for \textsc{score} questions, the expected score $\sum_k p_k v_k$. Several questions about one response (several
traits or marking points) share the state. Scoring questions are accepted only when $s$ contains the item and at least one of rubric,
reference answer or key points; we neither train nor serve answer-only scoring.

\subsection{Single forward pass with a pointer head}
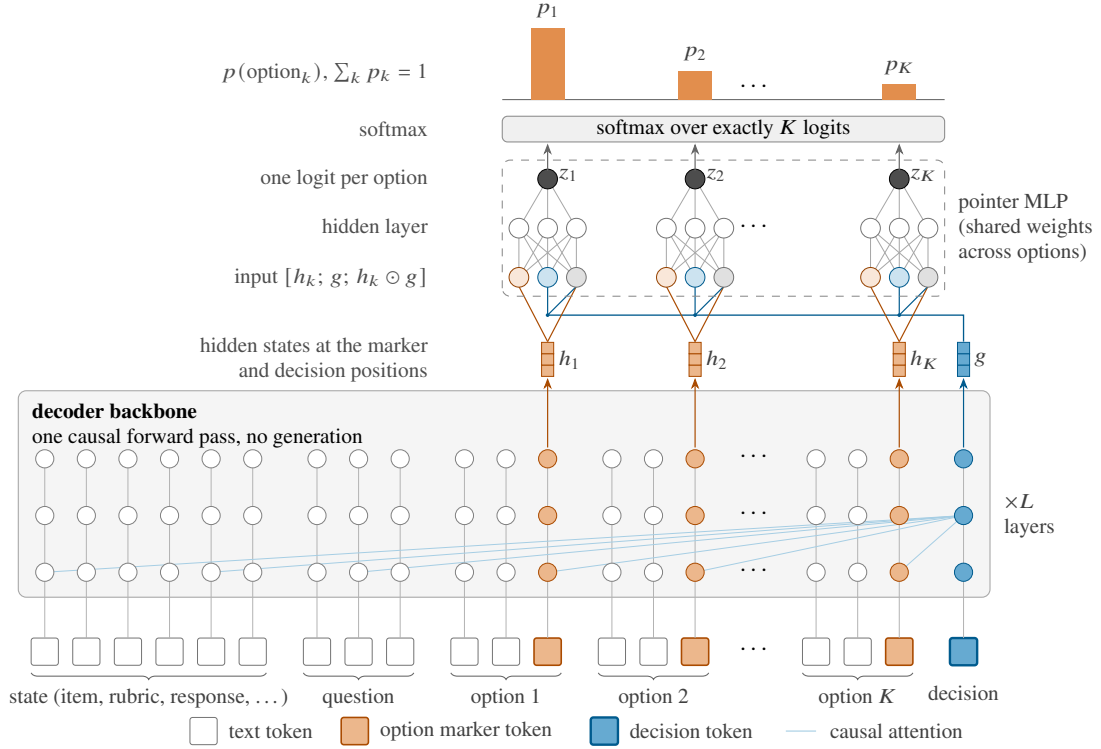
\begin{figure}[t]
\centering
\begin{tikzpicture}[x=1cm, y=1cm, font=\footnotesize,
  tok/.style={draw=black!55, line width=0.35pt, rounded corners=1pt, minimum size=3.6mm, inner sep=0pt, fill=white},
  mtok/.style={tok, draw=mkc!80!black, fill=mkc!45, line width=0.6pt},
  dtok/.style={tok, draw=dcc!80!black, fill=dcc!60, line width=0.9pt},
  neu/.style={circle, draw=black!50, line width=0.35pt, fill=white, minimum size=2.4mm, inner sep=0pt},
  mneu/.style={neu, draw=mkc!80!black, fill=mkc!45},
  dneu/.style={neu, draw=dcc!80!black, fill=dcc!60},
  pneu/.style={circle, draw=mkc!70!black, line width=0.35pt, fill=mkc!15, minimum size=2.6mm, inner sep=0pt},
  conn/.style={draw=black!30, line width=0.3pt},
  arr/.style={-{Stealth[length=1.5mm]}, line width=0.45pt},
  lab/.style={font=\scriptsize, text=black!75},
  rowlab/.style={font=\scriptsize, text=black!70, anchor=east, align=right}]
  \definecolor{mkc}{HTML}{D55E00}\definecolor{dcc}{HTML}{0072B2}
  \def\tokpos{0/0,0.55/0,1.1/0,1.65/0,2.2/0,2.75/0,3.6/0,4.15/0,4.7/0,5.55/0,6.1/0,6.65/1,7.5/0,8.05/0,8.6/1,10.2/0,10.75/0,11.3/1,12.15/2}
  \def\dotsx{9.4}
  \def\layerys{1.05,1.8,2.55}
  \fill[rounded corners=3pt, black!4] (-0.35,0.72) rectangle (12.5,3.45);
  \draw[rounded corners=3pt, black!35, line width=0.4pt] (-0.35,0.72) rectangle (12.5,3.45);
  \node[anchor=north west, align=left, font=\scriptsize] at (-0.3,3.42)
    {\textbf{decoder backbone}\\ one causal forward pass, no generation};
  \foreach \x/\t in \tokpos {
    \ifnum\t=1 \node[mtok] (t\x) at (\x,0) {};
    \else\ifnum\t=2 \node[dtok] (t\x) at (\x,0) {};
    \else \node[tok] (t\x) at (\x,0) {}; \fi\fi
  }
  \node at (\dotsx,0) {$\cdots$};
  \foreach \x/\t in \tokpos {
    \draw[conn] (\x,0.18) -- (\x,2.55);
  }
  \foreach \x in {0,2.2,4.15,6.65,8.6,11.3} { \draw[draw=dcc!35, line width=0.3pt] (\x,1.05) -- (12.15,1.8); }
  \foreach \y in \layerys {
    \foreach \x/\t in \tokpos {
      \ifnum\t=1 \node[mneu] at (\x,\y) {};
      \else\ifnum\t=2 \node[dneu] at (\x,\y) {};
      \else \node[neu] at (\x,\y) {}; \fi\fi
    }
    \node at (\dotsx,\y) {$\cdots$};
  }
  \node[lab, anchor=west, align=left] at (12.55,1.8) {$\times L$\\layers};
  \begin{scope}[decoration={brace, mirror, amplitude=2.5pt}, line width=0.4pt, black!60]
    \draw[decorate] (-0.18,-0.28) -- (2.93,-0.28) node[midway, below=2pt, lab] {state (item, rubric, response, \dots)};
    \draw[decorate] (3.42,-0.28) -- (4.88,-0.28) node[midway, below=2pt, lab] {question};
    \draw[decorate] (5.37,-0.28) -- (6.83,-0.28) node[midway, below=2pt, lab] {option 1};
    \draw[decorate] (7.32,-0.28) -- (8.78,-0.28) node[midway, below=2pt, lab] {option 2};
    \draw[decorate] (10.02,-0.28) -- (11.48,-0.28) node[midway, below=2pt, lab] {option $K$};
  \end{scope}
  \node[lab, below=1pt] at (12.15,-0.28) {decision};
  \node[tok] at (2.1,-1.05) {}; \node[lab, anchor=west] at (2.3,-1.05) {text token};
  \node[mtok] at (4.1,-1.05) {}; \node[lab, anchor=west] at (4.3,-1.05) {option marker token};
  \node[dtok] at (7.4,-1.05) {}; \node[lab, anchor=west] at (7.6,-1.05) {decision token};
  \draw[draw=dcc!40, line width=0.3pt] (9.8,-1.05) -- (10.2,-1.05); \node[lab, anchor=west] at (10.25,-1.05) {causal attention};
  \foreach \m/\k in {6.65/1, 8.6/2, 11.3/K} {
    \draw[arr, mkc!80!black] (\m,2.7) -- (\m,3.62);
    \foreach \j in {0,1,2} \draw[draw=mkc!80!black, fill=mkc!45, line width=0.35pt] (\m-0.08,3.64+0.15*\j) rectangle (\m+0.08,3.79+0.15*\j);
    \node[lab, anchor=west, inner sep=1pt] at (\m+0.1,3.86) {$h_{\k}$};
  }
  \draw[arr, dcc!80!black] (12.15,2.7) -- (12.15,3.62);
  \foreach \j in {0,1,2} \draw[draw=dcc!80!black, fill=dcc!60, line width=0.35pt] (12.07,3.64+0.15*\j) rectangle (12.23,3.79+0.15*\j);
  \node[lab, anchor=west, inner sep=1pt] at (12.25,3.86) {$g$};
  \node[rowlab] at (5.2,3.86) {hidden states at the marker\\ and decision positions};
  \draw[dcc!80!black, line width=0.5pt] (12.15,4.1) -- (12.15,4.45) -- (6.65,4.45);
  \foreach \m/\k in {6.65/1, 8.6/2, 11.3/K} {
    \foreach \a in {-0.38,0,0.38} { \foreach \b in {-0.38,0,0.38} \draw[conn, draw=black!35] (\m+\a,4.95) -- (\m+\b,5.6); \draw[conn, draw=black!35] (\m+\a,5.6) -- (\m,6.25); }
    \draw[mkc!80!black, line width=0.45pt] (\m,4.1) -- (\m-0.38,4.83);
    \draw[mkc!80!black, line width=0.45pt] (\m,4.1) -- (\m+0.38,4.83);
    \draw[dcc!80!black, line width=0.45pt] (\m,4.45) -- (\m,4.83);
    \draw[dcc!80!black, line width=0.45pt] (\m,4.45) -- (\m+0.38,4.83);
    \fill[dcc!80!black] (\m,4.45) circle (0.7pt);
    \node[pneu] at (\m-0.38,4.95) {}; \node[pneu, draw=dcc!80!black, fill=dcc!20] at (\m,4.95) {}; \node[pneu, fill=black!12, draw=black!55] at (\m+0.38,4.95) {};
    \foreach \a in {-0.38,0,0.38} \node[pneu, fill=white, draw=black!55] at (\m+\a,5.6) {};
    \node[pneu, fill=black!70, draw=black] (z\k) at (\m,6.25) {};
    \node[lab, anchor=west, inner sep=1pt] at (\m+0.1,6.3) {$z_{\k}$};
    \draw[arr, black!60] (\m,6.36) -- (\m,6.72);
  }
  \node at (\dotsx,5.6) {$\cdots$};
  \draw[dashed, black!45, rounded corners=3pt, line width=0.4pt] (6.05,4.7) rectangle (11.9,6.5);
  \node[lab, anchor=west, align=left] at (11.95,5.6) {pointer MLP\\ (shared weights\\ across options)};
  \node[rowlab] at (5.2,4.95) {input $[h_k;\,g;\,h_k\odot g]$};
  \node[rowlab] at (5.2,5.6) {hidden layer};
  \node[rowlab] at (5.2,6.25) {one logit per option};
  \draw[rounded corners=2pt, fill=black!6, draw=black!45, line width=0.4pt] (6.05,6.74) rectangle (11.9,7.08);
  \node[font=\scriptsize] at (8.975,6.91) {softmax over exactly $K$ logits};
  \node[rowlab] at (5.2,6.91) {softmax};
  \draw[black!60, line width=0.4pt] (6.05,7.3) -- (11.9,7.3);
  \foreach \m/\h/\k in {6.65/0.95/1, 8.6/0.38/2, 11.3/0.2/K} {
    \fill[mkc!70] (\m-0.22,7.3) rectangle (\m+0.22,7.3+\h);
    \node[lab, above=0pt] at (\m,7.3+\h) {$p_{\k}$};
  }
  \node at (\dotsx,7.45) {$\cdots$};
  \node[rowlab] at (5.2,7.65) {$p(\text{option}_k)$, $\sum_k p_k=1$};
\end{tikzpicture}
\caption{Single-pass decision model. The input lists the state, the question and the $K$ options; each option is followed by a marker
token and the input ends with a decision token. One causal forward pass of the decoder (no generation) gives hidden states $h_k$ at the
markers and $g$ at the decision token. A small pointer MLP, with weights shared across options, maps $[h_k;\,g;\,h_k\odot g]$ to one
logit per option, and a softmax over exactly these $K$ logits gives the answer distribution. Options cannot be invented or omitted.}
\label{r:fig:arch}
\end{figure}

The input is serialised as text in a fixed order: the state fields, the question, then each option followed by a dedicated marker
token, and finally a decision token (Figure~\ref{r:fig:arch}). One causal forward pass of the decoder gives hidden states
$h_1,\dots,h_K\in\mathbb{R}^d$ at the option markers and $g\in\mathbb{R}^d$ at the decision token. A two-layer MLP $\phi$, shared across
options, scores each option, and a softmax over exactly the $K$ logits of the question gives the output:
\begin{equation}
  z_k=\phi\big([\,h_k;\;g;\;h_k\odot g\,]\big),\qquad p_\theta(k\mid x)=\frac{\exp z_k}{\sum_{j=1}^{K}\exp z_j},\qquad k=1,\dots,K.
  \label{r:eq:pointer}
\end{equation}
The readout points at input positions, as in pointer networks~\citep{vinyals2015pointer}, and the language-model head is not used. Two
consequences follow. The output is typed by construction: the model can be wrong only by choosing badly, never by producing an
unparseable or out-of-range answer.
Each display order requires one prefill forward pass, without autoregressive answer generation. The number of passes does not increase with the number of options, but computational cost increases with the total input length, including the option text. Prefix caching can reuse the shared state across questions or display orders where their serialised prefixes coincide.

We chose this readout in controlled 0.8B comparisons (three seeds each, paired soft cross-entropy on held-out validation questions;
lower is better). A letter head that reads the language-model logits of option labels ``A/B/C'', the usual readout for
multiple-choice evaluation~\citep{hendrycks2021mmlu,robinson2023mcsb}, was significantly worse ($+0.033$) and agreed less often across
option orders (by 3.5 points), in line with the label and position biases of letter
readouts~\citep{zheng2024selectors,pezeshkpour2024order}. Of the other heads we tried, including set attention across the option
states~\citep{lee2019settransformer}, only mean pooling over the option span~\citep{reimers2019sbert} was significantly worse; the rest,
and starting from the instruction-tuned rather than the pre-trained base checkpoint, were within noise. We therefore start from the
base checkpoint.

\subsection{Soft-label training with a proper scoring rule}
The training loss is the soft cross-entropy
\begin{equation}
  \mathcal{L}(\theta)=\mathbb{E}_{(x,q)}\Big[-\sum_{k=1}^{K}q_k\log p_\theta(k\mid x)\Big],
  \label{r:eq:loss}
\end{equation}
a strictly proper scoring rule~\citep{gneiting2007proper}: for each $x$ its expectation is uniquely minimised by $p=q$. This strict-propriety property makes cross-entropy an appropriate objective for learning the target label distribution.
Where human raters disagree, $q$ preserves their observed disagreement. In our evaluated ablations, adding a Brier or ranked-probability
term for ordinal questions did not improve the reported metrics, so we retain cross-entropy alone.

\subsection{Order-averaged evaluation}
Predictions should be robust to arbitrary changes in option display order.
During training the option order and the surface
formatting of the input are randomised. At evaluation each question is presented under a set $\Pi(x)$ of display orders (on internal held-out
sets, four orders for \textsc{choice} and both directions for \textsc{yes/no} and \textsc{score}; on the benchmark suite, two orders per
question); the predictions are mapped back to the canonical order and averaged,
\begin{equation}
  \bar p(k\mid x)=\frac{1}{|\Pi(x)|}\sum_{\pi\in\Pi(x)}p_\theta\big(\pi(k)\mid \pi(x)\big),
  \label{r:eq:perm}
\end{equation}
and every metric is computed from $\bar p$. We also report \emph{order consistency}, the fraction of questions whose arg-max is the same
under every $\pi\in\Pi(x)$. Single-order evaluation can be misleading, because a model that has learned only position or label priors can
look strong in the original order; on our internal held-out sets every model is therefore also compared with a per-unit label-prior
baseline on the same rows.

\subsection{Model souping}
We save weights $\theta_t$ at one half and three quarters of a run of $T$ steps and at its end, and use their uniform average~\citep{wortsman2022soups} as the final model
\begin{equation}
  \bar\theta=\tfrac13\big(\theta_{T/2}+\theta_{3T/4}+\theta_{T}\big).
  \label{r:eq:avg}
\end{equation}
Averaging costs no training and, in our runs, trades at most a few tenths of a point of in-distribution accuracy for about one point of out-of-distribution accuracy (Section~\ref{r:analysis}).

\section{Training}
\label{r:training}

\subsection{Recipe}
All models are trained in a single supervised stage from Qwen3.5 base checkpoints~\citep{qwen35} at 0.8B, 2B and 4B. Qwen3.5 is a hybrid decoder in which most layers are linear-attention (Gated DeltaNet) layers and every fourth layer uses full attention; we use
only the text backbone. All parameters are trained (full fine-tuning), the pointer head with a larger learning rate than the backbone, and the learning rate decreases with model size. Training sequences pack several decision instances into one sequence with exact per-instance isolation, including the recurrent state of the linear-attention layers.

The learning-rate schedule is warmup-stable-decay (WSD)~\citep{hagele2024wsd}: a short linear warmup, a long constant phase and a final decay. One detail turned out to matter: the decay phase reads data the model has \emph{not} seen before, namely the tail of the epoch, rather than starting a second pass over seen data (Section~\ref{r:analysis}). Each model is trained for one epoch of the mixture, about 1.5B tokens or 4.6M decision instances, at a global batch of about 131k tokens per step; the 4B model uses fully sharded data parallelism. We use model souping for the final weights, Equation~\eqref{r:eq:avg}.

\subsection{Data}
\begin{figure}[h]
\centering
\includegraphics[width=0.8\linewidth]{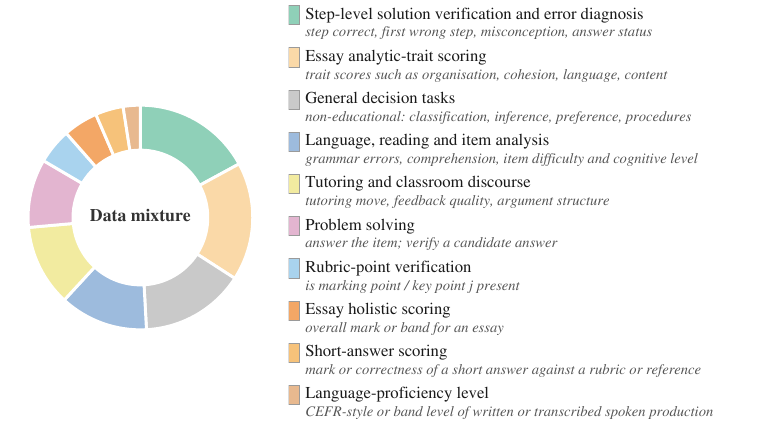}
\caption{Training data mixture by capability category (approximate relative composition by training tokens); slices and legend entries are ordered by size, and the grey sub-labels list typical decisions in each category.}
\label{r:fig:mix}
\end{figure}

The training mixture combines public educational and NLP datasets, programmatically generated tasks, and synthetic items and labels produced by third-party models (Section~\ref{r:statements}). The educational sources cover short-answer scoring, essay and learner-text scoring, step-level verification of worked solutions, student-error remediation and tutoring dialogues. Every source is converted to the typed-decision format of Section~\ref{r:approach} and assigned to a capability category. Figure~\ref{r:fig:mix} shows the approximate relative composition of the mixture. Scoring of free-text responses is split into short-answer scoring, essay holistic scoring, essay analytic-trait scoring and language-proficiency level; analytic-trait scoring is the largest of these, because one essay is scored on several traits. The data are mostly English and Chinese, with smaller amounts of other languages (for example German, Portuguese and learner texts in several European languages).

Within the educational categories, sources are balanced by a temperature over task families and capped per item so that no single question is repeated more than a few times per epoch. Labels come at several quality tiers (multi-rater human, single human, rule-exact, model-generated); all tiers are used with equal weight. Only rows whose context contains what a human marker would need are used for scoring decisions. Openly licensed sources include SciEntsBank and Beetle~\citep{dzikovska2013semeval}, SAF~\citep{filighera2022saf}, MERLIN~\citep{boyd2014merlin}, PRM800K~\citep{lightman2024prm}, MR-GSM8K~\citep{zeng2023mrgsm8k}, student-error remediation data~\citep{daheim2024stepverify}, MathDial~\citep{macina2023mathdial}, CIMA~\citep{stasaski2020cima} and MRBench~\citep{maurya2025mrbench} (CC BY, CC BY-SA or MIT licences); other sources are used under their respective terms (Section~\ref{r:statements}).

\subsection{Decontamination}
Training, validation and test splits are made at the level of content clusters. Items are grouped by exact match after text normalisation and by near-duplicate detection (MinHash signatures over character 5-grams with locality-sensitive hashing), and a cluster never crosses splits. All training data are then screened against every evaluation set used in this report, including the external benchmarks, by exact match, 8-gram overlap and MinHash near-duplicate detection on the item text, and matching training rows are removed. Source evaluation splits, and public test sets whose items circulate in other training corpora (for example grade-school and competition mathematics), are kept as block lists, so that copies inside other sources are caught by the same tests. Benchmarks that belong to task families represented in training are kept but reported separately as in-domain, and the problem-solving benchmark is reported on its subset of items with low n-gram overlap with the training data (Section~\ref{r:eval}).

\section{Evaluation}
\label{r:eval}

\subsection{Metrics}
The primary metric is accuracy: the arg-max of the order-averaged distribution $\bar p$ (Equation~\eqref{r:eq:perm}) against the gold
label (for multi-rater items, the official gold or the majority). For ordinal scoring tasks we also report quadratic weighted
kappa~\citep{cohen1968kappa}, computed per question or prompt and averaged. For $K$ ordered levels, with $O_{ij}$ the number of items
with gold level $i$ and predicted level $j$, $E_{ij}$ the count expected from the two marginal distributions, and
$w_{ij}=(i-j)^2/(K-1)^2$,
\begin{equation}
  \mathrm{QWK}=1-\frac{\sum_{i,j}w_{ij}O_{ij}}{\sum_{i,j}w_{ij}E_{ij}}.
  \label{r:eq:qwk}
\end{equation}
Calibration~\citep{guo2017calibration,li2023cue,li2026sedl}, which matters for routing uncertain cases to a
person~\citep{williamson2012framework}, is measured by the expected calibration error over 15 equal-width bins $B_1,\dots,B_{15}$ of the top-option probability,
\begin{equation}
  \mathrm{ECE}=\sum_{b=1}^{15}\frac{|B_b|}{N}\,\big|\mathrm{acc}(B_b)-\mathrm{conf}(B_b)\big|,
  \label{r:eq:ece}
\end{equation}
where $N$ is the number of questions, $\mathrm{acc}(B_b)$ the accuracy and $\mathrm{conf}(B_b)$ the mean top probability in bin $b$.
ECE and soft cross-entropy are secondary metrics and are reported raw, at temperature 1, unless stated otherwise: temperatures fitted on
in-distribution calibration data can hide over-confidence out of distribution. Confidence intervals are 95\% paired bootstrap intervals
that resample items within units (a unit is one dataset and question, e.g.\ one prompt and trait), with $B=1000$ resamples for the benchmark suite and $B=2000$ for comparisons on our internal held-out
sets (Section~\ref{r:res:scale}). Differences we describe as significant have paired 95\% bootstrap CIs excluding zero; we give point
estimates in the text and mark the others as within noise or not significant. The intervals reflect item sampling only, not
run-to-run variation (Section~\ref{r:limits}).

\subsection{Benchmark suite}
For comparison with other models we built a fixed suite of 20 sets, using only test or held-out splits, questions with at most 26
options, two display orders per question and unit-stratified samples of up to 1,500 questions. The sets are grouped by the ability
they test.
\begin{itemize}
  \item \emph{General decisions (5):} a hard subset of the public MMDM benchmark~\citep{mmdm2026} (MMDM-hard), the
        public split of JevBench~\citep{jevbench2026}, BoolQ~\citep{clark2019boolq}, ARC-Challenge~\citep{clark2018arc}, and a transfer set of held-out
        general decision datasets such as WANLI~\citep{liu2022wanli}.
  \item \emph{Reasoning (3):} a problem-solving set of science and maths items (from ARC-Easy~\citep{clark2018arc},
        OpenBookQA~\citep{mihaylov2018obqa}, high-school MMLU~\citep{hendrycks2021mmlu} and similar sources) restricted to low n-gram
        overlap with our training data, an in-family set of solving and answer-verification items, and held-out reading-comprehension questions from public
        datasets.
  \item \emph{Education, in-domain (8):} held-out items of the education task families in \edujev{}'s training data: short-answer
        scoring, essay scoring, rubric-point verification, step-level verification, error diagnosis, tutoring and classroom discourse,
        language assessment, and item analysis.
  \item \emph{Education, cross-domain (4):} unseen prompts of known education datasets, unseen education datasets of known task
        families, short-answer scoring on unseen prompts of a public dataset (QWK), and short essay scoring on unseen prompts of a
        public dataset (QWK).
\end{itemize}
For \edujev{}, the eight in-domain sets come from task families seen in training (with disjoint items), and the four cross-domain
sets test prompts or datasets never seen in training. Cells of open third-party models carry flags taken from their public documentation: \up{I} the benchmark belongs to a task
family in that model's training data, \up{P} partly in-distribution, \up{C} contaminated (rewritten test items in training). The hosted
APIs do not disclose their training data. Group averages are unweighted means of the benchmark accuracies; the education average
covers both in-domain and cross-domain sets.

\subsection{Baselines}
We compare with the open Jev-style decision models we could run through their own published inference code (OpenJev-4B~\citep{openjevdata}, JevK5-4B,
JevK5-2B, Tev1-4B, AstroHan-4B, eve-rlcd-0.6B and kotoba-DeBERTa-0.4B; minimal compatibility fixes only)\footnote{Hugging Face repositories:
\href{https://huggingface.co/shenjunhao/OpenJev-4B}{\texttt{shenjunhao/OpenJev-4B}},
\href{https://huggingface.co/alibiserikbay/JevK5}{\texttt{alibiserikbay/JevK5}},
\href{https://huggingface.co/alibiserikbay/JevK5-2B}{\texttt{alibiserikbay/JevK5-2B}},
\href{https://huggingface.co/togethercomputer/Tev1-4B-experimental}{\texttt{togethercomputer/Tev1-4B-experimental}},
\href{https://huggingface.co/AstroHan/decision-head-qwen3.5-4b-rlcd-32k}{\texttt{AstroHan/decision-head-qwen3.5-4b-rlcd-32k}},
\href{https://huggingface.co/anthonym21/qwen3-0.6b-rlcd-decision}{\texttt{anthonym21/qwen3-0.6b-rlcd-decision}} (eve-rlcd-0.6B),
\href{https://huggingface.co/com-kotobalabs/open-jev-deberta-v3-large}{\texttt{com-kotobalabs/open-jev-deberta-v3-large}}
(kotoba-DeBERTa-0.4B), and \href{https://huggingface.co/Qwen}{\texttt{Qwen/Qwen3.5-\{0.8B,2B,4B\}[-Base]}}.} and with zero-shot Qwen3.5
models at 0.8B--4B, read out through the next-token probabilities of option letters averaged over display orders. We also query five
commercial decision APIs through a routing service, using only endpoints with zero data retention. Section~\ref{r:statements} notes why
this comparison is not fully independent.

\section{Results}
\label{r:results}

All \edujev{} numbers are for the final weight-averaged checkpoints. Unless noted, each configuration is a single training run.

\subsection{Comparison with open decision models and hosted APIs}
\label{r:res:bench}

\begin{table}[!ht]
\centering\small
\setlength{\tabcolsep}{3.5pt}
\resizebox{\linewidth}{!}{%
\begin{tabular}{lrcccccccccc}
\toprule
 & & \multicolumn{6}{c}{general decisions} & \multicolumn{4}{c}{reasoning} \\
\cmidrule(lr){3-8}\cmidrule(lr){9-12}
model & size & MMDM-hard & JevBench & BoolQ & ARC-C & Transfer & \textbf{Avg} & solve (clean) & solve (family) & reading & \textbf{Avg} \\
\midrule
\multicolumn{12}{l}{\emph{Open Jev-style decision models}}\\
eve-rlcd-0.6B & 0.6B & 41.3 & 61.5 & 87.1\up{I} & 50.7 & 60.7\up{P} & 60.3 & 58.4 & 43.2 & 56.1 & 52.6 \\
JevK5-2B & 2B & 56.3 & 74.0 & 83.5 & 82.0 & 65.6 & 72.3 & 77.3 & 57.2 & 73.6 & 69.4 \\
AstroHan-4B & 4B & 72.0 & 76.6 & 86.9 & 92.5 & 76.2 & 80.8 & 84.5 & 68.0 & 82.0 & 78.2 \\
Tev1-4B & 4B & 72.9 & 75.3 & 91.8\up{I} & 92.7 & 72.9\up{P} & 81.1 & 84.9 & 68.4 & 79.9 & 77.7 \\
JevK5-4B & 4B & 73.2\up{P} & 87.9 & 87.8\up{I} & 93.8\up{I} & 75.9\up{I} & 83.7 & 87.4\up{P} & 70.4 & 84.1 & 80.6 \\
OpenJev-4B & 4B & 84.6\up{I} & 87.9\up{C} & 91.2\up{I} & 94.0\up{I} & 70.2 & 85.6 & 88.2\up{P} & 71.7 & 85.8 & 81.9 \\
\midrule
\multicolumn{12}{l}{\emph{Zero-shot Qwen3.5 (letter readout)}}\\
Qwen3.5-2B & 2B & 52.7 & 66.7 & 82.1 & 81.1 & 58.3 & 68.2 & 74.9 & 55.7 & 74.5 & 68.4 \\
Qwen3.5-4B & 4B & 64.5 & 76.2 & 88.3 & 91.9 & 69.3 & 78.0 & 85.5 & 68.0 & 79.9 & 77.8 \\
\midrule
\multicolumn{12}{l}{\emph{Hosted decision APIs}}\\
Jev 1.14 & -- & 86.8 & 88.3 & 93.0 & \textbf{97.8} & 78.6 & 88.9 & 93.0 & 81.6 & 89.1 & 87.9 \\
Microsoft Decision-1 & -- & 86.1 & 90.0 & 93.3 & 96.2 & 79.7 & 89.1 & 91.1 & 77.7 & 85.8 & 84.9 \\
Cloudflare Clef & -- & 81.5 & 87.4 & 92.4 & \textbf{97.8} & 78.3 & 87.5 & 90.7 & 74.3 & 89.1 & 84.7 \\
Drex Max & -- & \textbf{91.4} & \textbf{91.8} & \textbf{94.4} & 96.8 & 80.7 & \textbf{91.0} & \textbf{93.7} & 83.9 & \textbf{90.8} & \textbf{89.5} \\
pplx-decider v1.2 & -- & 90.0 & 89.6 & 93.7 & \textbf{97.8} & \textbf{81.6} & 90.5 & 92.8 & \textbf{84.0} & 89.1 & 88.6 \\
\midrule
\multicolumn{12}{l}{\emph{\edujev{}}}\\
\edujev{} 0.8B & 0.8B & 67.5 & 64.5 & 78.5 & 74.6 & 68.1 & 70.6 & 73.6 & 66.2 & 76.2 & 72.0 \\
\edujev{} 2B & 2B & 77.3 & 72.3 & 82.5 & 86.3 & 72.0 & 78.1 & 81.6 & 70.5 & 84.5 & 78.9 \\
\edujev{} 4B & 4B & 83.9 & 77.1 & 90.0 & 94.3 & 76.4 & 84.3 & 89.3 & 77.1 & \textbf{90.8} & 85.7 \\
\bottomrule
\end{tabular}}
\caption{General decision and reasoning benchmarks (accuracy, \%). Best in each column in bold (higher is better; ties all marked).
Flags on open third-party models follow their documentation: \up{I} in-domain family, \up{P} partly in-distribution, \up{C}
contaminated; the hosted APIs are queried zero-shot; their training data are undisclosed. Zero-shot Qwen3.5 models are read out
through option-letter probabilities. Reasoning: answering science and
maths items with low overlap with our training data, answering or verifying items of an in-family set, and reading comprehension.
Selected baselines are shown; all models, with 95\% CIs of the group averages, are in Table~\ref{r:tab:full}.}
\label{r:tab:general}
\end{table}

\begin{table}[t]
\centering\small
\setlength{\tabcolsep}{2.6pt}
\resizebox{\linewidth}{!}{%
\begin{tabular}{lcccccccc@{\hspace{9pt}}c@{\hspace{9pt}}cccc@{\hspace{9pt}}c@{\hspace{9pt}}c}
\toprule
 & \multicolumn{9}{c}{in-domain task families} & \multicolumn{5}{c}{cross-domain} & \\
\cmidrule(lr){2-10}\cmidrule(lr){11-15}
 & short ans. & essay & rubric & step & error & tutoring & language & item & & unseen & unseen & short-answer & short essay & & \textbf{Edu} \\
model & QWK & QWK & pts & verif. & diagn. & discourse & assess. & analysis & \textbf{Avg} & prompts & datasets & scoring QWK & scoring QWK & \textbf{Avg} & \textbf{Avg} \\
\midrule
\multicolumn{16}{l}{\emph{Open Jev-style decision models}}\\
AstroHan-4B & 45.7 & 32.1 & 34.5 & 63.8 & 52.8 & 55.0 & 48.3 & 55.1 & 48.6 & 36.3 & 63.5 & 37.6 & 8.1 & 38.4 & 45.2 \\
Tev1-4B & 61.2 & 29.9 & 34.1 & 61.8$^\dagger$ & 50.6 & 51.2 & 50.8$^\dagger$ & 53.2 & 48.5 & 40.7 & 65.1 & 46.1 & 30.1 & 46.9 & 48.0 \\
JevK5-4B & 51.7 & 27.3 & 40.0 & 68.2 & 52.3 & 53.3 & 41.2 & 56.9 & 50.0 & 37.8 & 58.7 & 50.0 & 16.1 & 46.4 & 48.8 \\
OpenJev-4B & 40.3 & 26.4 & 42.0 & 63.8 & 53.3 & 49.7 & 46.7 & 57.6 & 49.4 & 38.9 & 63.1 & 41.2 & 44.3 & 47.8 & 48.9 \\
\midrule
\multicolumn{16}{l}{\emph{Zero-shot Qwen3.5 (letter readout)}}\\
Qwen3.5-4B & 50.3 & 22.4 & 40.7 & 61.2 & 51.9 & 54.2 & 43.0 & 56.7 & 48.4 & 35.3 & 56.3 & \textbf{54.5} & 2.3 & 40.2 & 45.7 \\
\midrule
\multicolumn{16}{l}{\emph{Hosted decision APIs}}\\
Jev 1.14 & 68.6 & 36.2 & 44.3 & 69.6 & 56.8 & 59.4 & 54.1 & 59.7 & 55.5 & 40.7 & 68.7 & 50.1 & 9.5 & 42.1 & 51.0 \\
Microsoft Decision-1 & 63.0 & 34.7 & 38.7 & 71.8 & 57.5 & 60.5 & 55.8 & 59.3 & 53.5 & 36.2 & 69.8 & 52.5 & 5.3 & 40.6 & 49.2 \\
Cloudflare Clef & 67.5 & 38.3 & 50.2 & 74.4 & 58.5 & 63.2 & 54.8 & 60.2 & 57.8 & 41.8 & 71.1 & 48.4 & 21.5 & 47.7 & 54.4 \\
Drex Max & 71.3 & 44.2 & 50.2 & 81.1 & 59.5 & 64.5 & 61.1 & 60.7 & 60.1 & 46.9 & \textbf{72.3} & 43.0 & 43.9 & 53.4 & 57.9 \\
pplx-decider v1.2 & 66.7 & 43.3 & 51.0 & 81.0 & 60.0 & 64.4 & 62.6 & 59.5 & 59.6 & 47.9 & 71.6 & 38.4 & \textbf{58.2} & \textbf{55.4} & 58.2 \\
\midrule
\multicolumn{16}{l}{\emph{\edujev{}}}\\
\edujev{} 0.8B & 69.6 & 59.6 & 59.5 & 81.5 & 77.5 & 75.4 & 60.9 & 79.7 & 70.3 & 59.8 & 58.7 & 37.7 & 38.7 & 47.9 & 62.8 \\
\edujev{} 2B & 71.5 & 60.2 & \textbf{60.2} & 84.1 & 76.8 & 77.2 & 64.7 & 82.0 & 71.8 & 59.4 & 61.1 & 41.4 & 30.7 & 46.0 & 63.2 \\
\edujev{} 4B & \textbf{74.0} & \textbf{62.3} & \textbf{60.2} & \textbf{88.2} & \textbf{78.8} & \textbf{78.7} & \textbf{68.0} & \textbf{82.6} & \textbf{73.7} & \textbf{60.7} & 66.8 & 51.8 & 43.5 & 55.2 & \textbf{67.6} \\
\bottomrule
\end{tabular}}
\caption{Education benchmarks (\%; QWK $\times100$). In-domain: held-out items of the eight education task families in \edujev{}'s
training data. Cross-domain: unseen prompts of known datasets, unseen datasets of known families, and short-answer and short essay
scoring on unseen prompts of two public datasets. The four scoring columns marked QWK (two in-domain, two cross-domain) show QWK; all other columns are accuracies. Every average
uses the \emph{accuracy} of all its sets, including the scoring sets, so the averages cannot be recomputed from the QWK columns; Edu
Avg is the mean over all twelve sets. Best in each column in bold; $^\dagger$ partial coverage: some items exceeded a model's context
window and were not scored. Selected baselines are shown; all models are in Table~\ref{r:tab:full}.}
\label{r:tab:edu}
\end{table}

Tables~\ref{r:tab:general} and~\ref{r:tab:edu} and Figure~\ref{r:fig:scatter} summarise the comparison. All differences below are
paired, for \edujev{} 4B unless stated, in accuracy points (kappa points for QWK), and significant unless marked otherwise.

On education, \edujev{} 4B has the highest average of all models (67.6), followed by our 2B (63.2) and 0.8B (62.8) models, so that
every \edujev{} model is above every other model in the comparison. The 4B model outperforms the strongest API, pplx-decider v1.2, by 9.4
points (67.6 vs.\ 58.2) and the other four APIs by 9.7 (Drex Max) to 18.3 points (Microsoft Decision-1), and it outperforms the open 4B models by 18.7 (OpenJev-4B) to
22.4 points (AstroHan-4B). Most of the lead comes from the eight in-domain families, where it is 13.6 (Drex Max) to 20.2 points (Microsoft
Decision-1) over the APIs and 23.8 to 25.2 points over the open 4B models. The 4B model is best on seven of the eight in-domain sets,
including families beyond response scoring such as error diagnosis, tutoring discourse and item analysis, and joint best (with our 2B
model) on rubric-point verification.

On the four cross-domain education sets, the \edujev{} 4B model achieves 55.2\% average accuracy, compared with 55.4\% for pplx-decider v1.2. The paired difference of $-0.2$ percentage points is not statistically significant.
\edujev{} 4B outperforms the
other four APIs, by 1.8 (Drex Max) to 14.6 points (Microsoft Decision-1), and the open 4B models, by 7.4 to 16.8 points. It is far ahead on unseen prompts
of known datasets (60.7 vs.\ at most 40.7 for open models and 47.9 for the APIs), and on short-answer scoring its QWK of 51.8 (41.4 at
2B) ranks third, behind zero-shot Qwen3.5-4B (54.5) and Microsoft Decision-1 (52.5), both differences within noise. It trails four of
the five APIs on unseen datasets (66.8 vs.\ 69.8 to 72.3) and is level with Jev 1.14 there (68.7, not significant), and it trails
pplx-decider v1.2 on short essay scoring (QWK 43.5 vs.\ 58.2), where it is level with OpenJev-4B (44.3) and Drex Max (43.9). The 2B model (46.0) is level with the strongest open 4B
models on these sets ($-1.9$ against OpenJev-4B, not significant).

On reasoning, the 4B model (85.7) outperforms the open 4B models by 3.8 (OpenJev-4B) to 8.0 points (Tev1-4B) and is level with two
APIs, Microsoft Decision-1 ($+0.9$) and Cloudflare Clef ($+1.0$), both within noise; it trails the other three APIs by 2.2 to 3.7
points. On general decisions, the 4B model (84.3) is level with JevK5-4B ($+0.6$, not significant), trails OpenJev-4B by 1.3 points,
outperforms the zero-shot Qwen3.5-4B model by 6.3 points, and trails the five APIs by 3.2 to 6.7 points. JevBench is its largest gap to
the open models (77.1 vs.\ 87.9 for the two strongest 4B models, one of which is flagged as contaminated).

These results show that \edujev{} leads every compared model on education tasks of the kinds it was trained for, and that at 4B it
carries much of that lead to unseen prompts and matches the strongest API on the cross-domain sets, while general decisions and reasoning, where
broad pre-training matters more, remain the strength of the hosted APIs.

\subsection{Scaling with model size}
\label{r:res:scale}
\begin{table}[t]
\centering\small
\setlength{\tabcolsep}{4pt}
\begin{tabular}{lccccc}
\toprule
model & in-dist.\ acc & OOD acc & solve / maths & general avg & education avg \\
\midrule
\edujev{} 0.8B & 75.0 & 62.1 & 75.1 / 46.8 & 70.6 & 62.8 \\
\edujev{} 2B & 76.5 & 64.1 & 83.5 / 52.5 & 78.1 & 63.2 \\
\edujev{} 4B & \textbf{78.0} & \textbf{67.2} & \textbf{90.2} / \textbf{61.2} & \textbf{84.3} & \textbf{67.6} \\
\bottomrule
\end{tabular}
\caption{Final \edujev{} models by size. In-distribution, OOD and solve / maths are measured on our internal held-out sets, not on
the benchmark sets of Table~\ref{r:tab:general}: in-distribution is held-out validation questions of the training task families; OOD is
the mean of the internal transfer, unseen-prompt and unseen-dataset sets; solve is answering science and maths items, and maths its
maths part. General and education averages are the benchmark averages of Tables~\ref{r:tab:general} and~\ref{r:tab:edu}. Best in each column in bold. Matched-data paired gains are
given in the text.}
\label{r:tab:scale}
\end{table}

Table~\ref{r:tab:scale} lists the final models.
All aggregate metrics shown in the table improve with model size, although improvements vary across measures and are not monotonic on every individual benchmark.
To separate size from data, we
compare models trained on identical data, schedule and token budget: a 2B model trained on the 0.8B model's data for the first step,
and the final 2B and 4B models, which share their data, for the second. Gains are paired and significant, except that solve and
maths (small internal sets) are reported as point estimates only. From 0.8B to 2B the gain is small on in-distribution
questions ($+1.3$), out of distribution ($+1.6$) and on the education average ($+0.8$), and larger on general decisions ($+6.4$) and
on solving and maths ($+9.1$ each). From 2B to 4B the in-distribution gain is again small ($+1.5$), while out-of-distribution accuracy
($+3.1$), the education average ($+4.4$), general decisions ($+6.2$) and reasoning ($+6.9$) improve more; the education gain comes
mostly from the cross-domain sets ($+9.3$, against $+1.9$ on the in-domain families). With three model sizes, we do not fit a scaling law.
In these experiments, the increase from 2B to 4B produces larger gains on several transfer and general-reasoning measures than on in-domain assessment accuracy. The smaller in-domain gains are consistent with a data bottleneck; separating data and capacity limitations requires varying training-data volume at each model size.

Before each larger run we registered predictions with 80\% intervals. For 0.8B to 2B, all endpoints fell inside them. For 2B to 4B,
the change in validation soft cross-entropy ($-0.045$, predicted $-0.043$ with interval $-0.055$ to $-0.031$) and the held-out scoring
QWK fell inside, and in-distribution accuracy was at the upper end, while the out-of-distribution, general and education gains were
above their intervals: size helped beyond the training distribution more than we expected.

\subsection{Serving efficiency}
\label{r:res:serve}
Because no tokens are generated, serving cost is one prefill per question and display order. We provide \edujev{} through a model plugin for
vLLM~\citep{kwon2023vllm} that applies the pointer head as a pooling step, and through a direct PyTorch path; both reuse the training
serialiser so that inputs are token-identical to training. On one H100 GPU, through the vLLM plugin, a 2B model processes about 150 decisions per second (corresponding to about
110k prompt tokens per second) on a realistic mix of request lengths at batch size 256, against about 120 per second for the direct path; for
small batches the direct path is faster. Prefix caching gives a further 2.5--3.8$\times$ when many questions share a long state, such as
several marking points of one response. Outputs match offline evaluation within the bf16 noise band: arg-max agreement with an fp32
reference is 99.3--99.6\%, the same as the agreement between bf16 and fp32 inference in the reference implementation itself.

\section{Analysis}
\label{r:analysis}

The recipe of Section~\ref{r:training} was settled in controlled runs, mostly at 0.8B and 2B. This section summarises what mattered,
what did not, and how well the output probabilities can be trusted.

\subsection{What mattered}
Data coverage gave the largest improvements, more than any change of architecture or objective. Problem-solving data (answering and
verifying items) are a clear example: in a small-budget ablation, increasing their share improved maths solving by 13 points and answer
verification by 6, without hurting in-distribution scoring. Before such data were added, judging a candidate answer without a reference
was bounded by the model's own ability to solve the item. More educational data also kept helping: at 2B, one full epoch raised
in-distribution accuracy from 70.7\% to 76.2\% (final weights of a development run, before souping) over a run with a sixth of the tokens.

Running the learning-rate decay on unseen data mattered for generalisation. When the decay instead started a second pass over data the
model had already seen, in-distribution metrics improved but raw out-of-distribution cross-entropy rose from 0.76 to 0.86 during the
decay: confidence out of distribution grew while accuracy did not, and temperatures fitted on in-distribution data hid this. Decaying on
the unseen tail of the epoch, from the same checkpoint and for the same number of steps, lowered raw out-of-distribution cross-entropy
by 0.051 with in-distribution quality unchanged.

Model souping is a small but consistent trade. With the decay on unseen data the final weights are already good, and averaging the
half, three-quarter and final checkpoints moves in-distribution accuracy by at most 0.3 points in either direction while adding 0.6 to
1.4 points out of distribution; at 4B the average is 0.5 points better out of distribution and 0.2 points lower in distribution than
the final weights. Interpolating the backbone back towards its pre-trained weights~\citep{wortsman2022wiseft} was dominated by
model souping.

\subsection{Calibration}
Raw (temperature 1) calibration is good where the model has seen the task family and degrades outside it. On the benchmark suite the
ECE of \edujev{} 4B averages 0.043 on general decisions, 0.056 on reasoning and 0.044 on the in-domain education sets, lower than every
hosted API on the in-domain sets (0.086 to 0.151) and lower than the best-calibrated open 4B model (JevK5-4B, 0.085). On the cross-domain education
sets it rises to 0.161: within the range of the APIs (0.152 to 0.251) but worse than that model (0.110). The internal held-out sets show
the same gradient, from 0.016 in distribution to 0.062 on transfer, 0.066 on unseen prompts and 0.125 on unseen datasets. Predictions
are also stable under re-ordering: for the 2B model, the mean total-variation distance between the distributions for different display
orders is 0.017 in distribution and 0.023 to 0.026 out of distribution, and 97\% of in-distribution questions have the same arg-max in
every order. These results suggest that the confidence can be used for routing within known task families, but should be re-checked on
each new assessment.

\subsection{What did not help}
Reinforcement learning on the pointer readout, with actions sampled from the output distribution and rewarded by agreement with the
label, did not beat continued supervised training in more than thirty controlled second-stage runs at 2B, across binary, ordinal and proper
rewards and the usual regularisers. Binary-reward RL raised the mean top probability on validation from 0.80 to 0.98 while reducing accuracy by 0.8 to
1.2 percentage points. The reason is structural: for a single readout trained on labelled data, the expected policy gradient under a
label-derived reward is a reweighted cross-entropy gradient (with a binary reward, $p_{\text{gold}}$ times it), so such rewards add no
information beyond the labels, and improper rewards push the output towards one-hot. This is consistent with the analysis of \citet{dft2025}, who show that the supervised fine-tuning gradient
is itself a policy gradient with an implicit reward, and with \citet{openjevrlcd2026}, who prove that reinforcement learning for
calibrated decisions cannot beat cross-entropy when the uncertainty comes from annotator disagreement; where reinforcement learning
does improve classification, it acts on generated reasoning before the answer~\citep{gta2025,openjevrlcd2026}. We therefore keep a purely supervised recipe and reserve RL for generated reasoning
(Section~\ref{r:conclusion}).

Training only the pointer head on a frozen backbone underfits badly (in-distribution accuracy 48\% against 70\% for full fine-tuning
on the same budget), and larger heads or higher head learning rates do not close the gap, so full fine-tuning is required at these
sizes. An abstention option trained on irrelevant option sets and richer pointer heads were neutral or worse.

\section{Limitations}
\label{r:limits}
Alongside these results, the current models and evaluation have several limitations. First, generalisation to new assessments is
incomplete. The strongest results are on task families represented in training, and the 4B model trails four of the five APIs on unseen datasets (66.8
vs.\ 69.8 to 72.3) and the strongest API on short essay scoring on unseen prompts (QWK 43.5 vs.\ 58.2); calibration is also weaker outside the
training families. Users should therefore check agreement and calibration on a sample of marked responses for each new assessment, and
adapt the model or route low-confidence cases to a teacher where needed. Second, eight of the twelve education benchmarks are held-out
items of task families we train on, which, as far as their documentation discloses, the compared models were not specifically trained
for, so part of the education lead reflects targeted
training data; we report the in-domain and cross-domain groups separately for this reason and plan to extend the cross-domain sets.
Third, \edujev{} trails the hosted APIs on general decisions and three of them on reasoning, so it is best used as an assessment
specialist rather than a general decision model; broader general-decision data are a natural next step. Fourth, each configuration is a
single training run, and bootstrap intervals over items do not capture run-to-run variation, so small single-run differences on small
benchmarks should be read with caution; multi-seed replicates of the main comparisons are planned. Finally, the models are text only
(handwriting must be transcribed), are trained mostly on English and Chinese data, and have not been evaluated in a teacher-in-the-loop deployment;
such a study, including an audit of differences across student populations, is required before operational use.

\section{Conclusion and future work}
\label{r:conclusion}

\edujev{} demonstrates that education-focused supervised training can produce small models that return structurally valid assessment decisions without autoregressive answer generation. Across twelve education evaluation sets, the 4B model achieves the highest average accuracy among the evaluated models, with its largest gains on task families represented in training. Its average accuracy across four cross-domain education sets is close to that of the strongest commercial decision API. The main priorities are  to strengthen transfer, establish the reliability of confidence-based referral, and isolate the contributions of the training mixture and pointer readout through matched-data comparisons.

One direction is to let the model reason before making its decisions. An optional reasoning stage could generate a rationale before the answer options are processed, allowing the same rationale to be reused across display orders. With a reasoning budget of zero, the model would reduce to the current single-pass design. Reinforcement learning could optimise this stage by rewarding rationales that improve the pointer’s predictions under a proper scoring rule. Applying reinforcement learning directly to the readout did not improve performance in our experiments. Teacher-generated marking-point annotations could provide initial supervision for rationale generation.
A second direction is to strengthen data and evaluation. Scaling from 2B to 4B produced the largest gains where
the 2B model was weakest, with more modest improvements on in-domain task families.
This pattern motivates investigating whether broader coverage and higher-quality supervision can improve performance on these families.
Multi-seed replicates and evaluation on further assessment datasets and essay prompts are also needed to establish the robustness of the main comparisons.

\section{Statements}
\label{r:statements}

\subsection*{Intended use}
\edujev{} is a research model for assisting assessment: marking short and extended responses against a given rubric, checking marking
points, verifying solution steps, and flagging uncertain cases for human review. It is meant to be used with the item and its rubric,
reference answer or key points in the context, and our serving code rejects scoring requests without them. It is not intended for
high-stakes decisions about individual students without human oversight. Calibrated confidence is meant to support routing uncertain
responses to a teacher, and should be re-checked on each new assessment before it is relied on. Automated scoring can be biased across
student populations, languages and writing styles; we have not audited such effects.

\subsection*{Data and ethics}
The training mixture combines public educational and NLP datasets (about 44\% of training tokens under permissive or share-alike licences, and some without a stated licence), programmatically generated tasks, and synthetic items and labels produced by third-party models, including a commercial decision API. We excluded sources whose terms we read as forbidding model training; for third-party model outputs, terms of use may restrict downstream use, which is one reason the models are intended for research only. The public evaluation benchmarks (MMDM, JevBench, BoolQ, ARC, OpenBookQA, MMLU, WANLI) are used under their MIT, CC BY-SA, CC BY or Apache-2.0 licences.

\subsection*{Independence of the comparison}
The comparison with open Jev-style models and hosted APIs is not fully independent: our general decision data include publicly released typed-decision training data of the kind some of these models were trained on, as well as synthetic labels from third-party decision models; conversely, several open models and all hosted APIs do not disclose their training data.

\subsection*{AI assistance}
Code, experiments and drafts of this report were produced with the assistance of an AI coding agent working under the direction of the authors, who designed the studies, set the decision rules and reviewed the results. All numbers are taken from logged experiment outputs.

\section*{Acknowledgements}
This work used King's Computational Research, Engineering and Technology Environment
(CREATE)~\citep{kcl2022create}.

\bibliographystyle{plainnat}
\bibliography{references}

\clearpage
\appendix
\section{Full benchmark results}
\label{r:app:bench}

Table~\ref{r:tab:full} gives the group averages of every model in the comparison with their 95\% bootstrap CIs, which resample items
within units ($B=1000$); per-benchmark results for selected models are in Tables~\ref{r:tab:general} and~\ref{r:tab:edu}. Each question is shown in two
display orders and the predictions are averaged. Every model, including the hosted APIs, is scored on the same items. Third-party models
were run through their own published inference code with minimal compatibility fixes (dependency checks, maximum question length,
truncation of over-long states); a few inputs beyond a model's context limit were truncated in the middle of the state.

\begin{table}[!ht]
\centering\small
\setlength{\tabcolsep}{3pt}
\resizebox{\linewidth}{!}{%
\begin{tabular}{lrccccc}
\toprule
model & size & general (5) & reasoning (3) & education in-domain (8) & education cross-domain (4) & education (12) \\
\midrule
\multicolumn{7}{l}{\emph{Open Jev-style decision models}}\\
kotoba-DeBERTa-0.4B & 0.4B & 58.9 [57.5, 60.5] & 49.3 [47.2, 51.5] & 34.8 [33.8, 35.8] & 31.7 [30.5, 33.1] & 33.8 [32.9, 34.6] \\
eve-rlcd-0.6B & 0.6B & 60.3 [58.8, 61.7] & 52.6 [50.2, 54.8] & 37.0 [35.9, 38.0] & 34.5 [33.1, 35.9] & 36.1 [35.3, 37.0] \\
JevK5-2B & 2B & 72.3 [70.8, 73.8] & 69.4 [67.4, 71.4] & 40.0 [38.9, 41.1] & 36.6 [35.1, 38.2] & 38.9 [38.0, 39.8] \\
AstroHan-4B & 4B & 80.8 [79.4, 82.0] & 78.2 [76.4, 79.8] & 48.6 [47.5, 49.8] & 38.4 [37.2, 39.8] & 45.2 [44.3, 46.1] \\
Tev1-4B & 4B & 81.1 [79.8, 82.3] & 77.7 [76.0, 79.5] & 48.5 [47.3, 49.6] & 46.9 [45.4, 48.4] & 48.0 [47.1, 48.9] \\
JevK5-4B & 4B & 83.7 [82.5, 84.8] & 80.6 [79.0, 82.3] & 50.0 [48.9, 51.1] & 46.4 [45.0, 47.9] & 48.8 [47.9, 49.7] \\
OpenJev-4B & 4B & 85.6 [84.5, 86.5] & 81.9 [80.3, 83.3] & 49.4 [48.2, 50.5] & 47.8 [46.3, 49.3] & 48.9 [48.0, 49.8] \\
\midrule
\multicolumn{7}{l}{\emph{Zero-shot Qwen3.5}}\\
Qwen3.5-0.8B & 0.8B & 61.1 [59.5, 62.7] & 59.6 [57.5, 61.7] & 37.4 [36.2, 38.5] & 40.2 [38.7, 41.6] & 38.3 [37.4, 39.2] \\
Qwen3.5-0.8B-Base & 0.8B & 59.5 [57.9, 61.2] & 62.3 [60.3, 64.3] & 35.4 [34.2, 36.4] & 27.5 [26.3, 28.6] & 32.7 [31.9, 33.5] \\
Qwen3.5-2B & 2B & 68.2 [66.7, 69.6] & 68.4 [66.3, 70.4] & 39.3 [38.2, 40.5] & 36.1 [34.7, 37.5] & 38.2 [37.3, 39.1] \\
Qwen3.5-2B-Base & 2B & 67.6 [66.2, 69.0] & 69.2 [67.2, 71.2] & 38.5 [37.3, 39.6] & 40.8 [39.3, 42.3] & 39.3 [38.2, 40.1] \\
Qwen3.5-4B & 4B & 78.0 [76.7, 79.3] & 77.8 [76.0, 79.5] & 48.4 [47.3, 49.6] & 40.2 [38.9, 41.6] & 45.7 [44.8, 46.5] \\
Qwen3.5-4B-Base & 4B & 76.7 [75.3, 77.9] & 78.8 [77.1, 80.6] & 48.8 [47.6, 50.0] & 43.5 [42.0, 44.9] & 47.0 [46.1, 47.9] \\
\midrule
\multicolumn{7}{l}{\emph{Hosted decision APIs}}\\
Jev 1.14 & -- & 88.9 [87.8, 89.8] & 87.9 [86.5, 89.2] & 55.5 [54.4, 56.6] & 42.1 [40.7, 43.3] & 51.0 [50.2, 51.9] \\
Microsoft Decision-1 & -- & 89.1 [88.1, 90.0] & 84.9 [83.3, 86.5] & 53.5 [52.4, 54.6] & 40.6 [39.4, 41.9] & 49.2 [48.4, 50.1] \\
Cloudflare Clef & -- & 87.5 [86.3, 88.4] & 84.7 [83.3, 86.1] & 57.8 [56.7, 58.9] & 47.7 [46.2, 49.1] & 54.4 [53.5, 55.3] \\
Drex Max & -- & \textbf{91.0} [90.1, 91.9] & \textbf{89.5} [88.2, 90.7] & 60.1 [59.0, 61.2] & 53.4 [52.0, 54.8] & 57.9 [57.0, 58.7] \\
pplx-decider v1.2 & -- & 90.5 [89.5, 91.4] & 88.6 [87.3, 90.0] & 59.6 [58.6, 60.7] & \textbf{55.4} [54.0, 56.8] & 58.2 [57.3, 59.1] \\
\midrule
\multicolumn{7}{l}{\emph{\edujev{}}}\\
\edujev{} 0.8B & 0.8B & 70.6 [69.2, 72.1] & 72.0 [69.9, 73.9] & 70.3 [69.2, 71.3] & 47.9 [46.4, 49.3] & 62.8 [62.0, 63.7] \\
\edujev{} 2B & 2B & 78.1 [76.7, 79.4] & 78.9 [77.1, 80.6] & 71.8 [70.8, 72.9] & 46.0 [44.6, 47.4] & 63.2 [62.4, 64.0] \\
\edujev{} 4B & 4B & 84.3 [83.2, 85.5] & 85.7 [84.3, 87.1] & \textbf{73.7} [72.7, 74.8] & 55.2 [53.7, 56.7] & \textbf{67.6} [66.8, 68.4] \\
\bottomrule
\end{tabular}}
\caption{Group averages (accuracy, \%) with 95\% bootstrap CIs in brackets. Best in each column in bold. The in-domain education sets
are held-out items of task families in \edujev{}'s training data.}
\label{r:tab:full}
\end{table}

\section{Evaluation protocol details}
\label{r:app:protocol}
On the internal held-out sets, choice questions are evaluated in four distinct orders (original, reversed and two seeded random orders;
fewer when $K$ is small), yes/no questions with either answer first, and score questions in ascending and descending order. Option
labels, when shown, are attached after permutation.
For each question, we map predictions from every display order back to canonical option identities and average the probability distributions. Accuracy, cross-entropy and calibration metrics are then computed from this averaged distribution, and QWK is computed from the resulting predicted categories across questions. We do not average the separately computed metrics of the display-order variants.

Every model is also compared on these sets with a label-prior baseline: the mean training label of the same item or dataset, per option
identity. A model that has learned only these priors can look strong in the original order, so development decisions were made on
paired differences against this baseline.

Micro averages weight questions; macro averages weight units (dataset $\times$ question) within families and then families. On the
benchmark suite, QWK is computed per unit with at least 15 items and then averaged. Comparisons on the internal held-out sets use a paired
bootstrap over units and items within units ($B=2000$); benchmark-suite tables and differences use $B=1000$. Where a decision rule or a predicted effect was fixed before a run (model size, learning
rate, reinforcement-learning variants), it is reported as such, and deviations are stated in the text. The final checkpoint (the
average of the half, three-quarter and final weights) was chosen by a rule fixed in advance, which combined in-distribution
accuracy, out-of-distribution accuracy and a held-out scoring QWK.

\end{document}